\documentclass[letterpaper, 10 pt, conference]{ieeeconf}  

\usepackage{algorithm}
\usepackage{algpseudocode}
\usepackage{amsmath}
\usepackage{amssymb}   
\usepackage{bm}        
\usepackage{booktabs}
\usepackage{xcolor}
\usepackage{nicefrac}       
\usepackage{microtype}      
\usepackage{mathtools}
\usepackage{multirow}
\usepackage{hyperref}
\usepackage{float}
\usepackage{caption}
\usepackage{graphicx}
\usepackage{cuted} 
\let\labelindent\relax
\usepackage{enumitem}

\usepackage{todonotes}

\newcommand{\ourwebsite}{\href{umi-on-air.github.io}{umi-on-air.github.io}}
\definecolor{custom_blue_1}{rgb}{0.263,0.388,0.847}
\definecolor{custom_orange_1}{rgb}{1.00, 0.541, 0.}
\definecolor{custom_green_1}{rgb}{0.235, 0.706, 0.294}
\definecolor{custom_yellow_1}{rgb}{0.9, 0.78, 0.31}
	
\IEEEoverridecommandlockouts                              

\title{\LARGE \bf
UMI-on-Air: Embodiment-Aware Guidance \\ for Embodiment-Agnostic Visuomotor Policies
\vspace{-1mm}
}

\author{Harsh Gupta$^{\dagger}$\quad Xiaofeng Guo$^{\dagger}$\quad Huy Ha$^{\ddagger}$\quad Chuer Pan$^{\ddagger}$\quad Muqing Cao$^{\dagger}$\quad  Dongjae Lee$^{\dagger}$\quad  \\ Sebastian Scherer$^{\dagger}$\quad Shuran Song$^{\ddagger}$\quad Guanya Shi$^{\dagger}$
\thanks{$^{\dagger}$Carnegie Mellon University\quad $^{\ddagger}$Stanford University}}

\begin{document}

\maketitle

\thispagestyle{empty}
\pagestyle{empty}

\begin{strip}
\vspace{-16 mm}
    \centering
    \includegraphics[width=\textwidth]{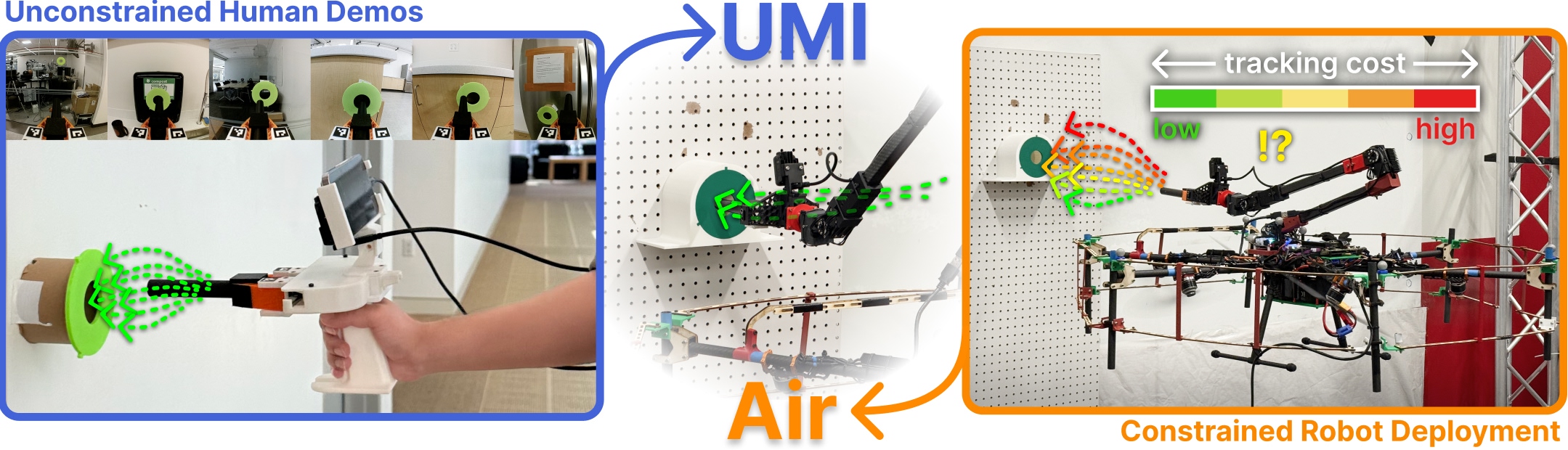}
    \vspace{-7mm}
        \captionof{figure}{\textbf{UMI-on-Air with Embodiment-Aware Guidance.} Standard UMI (Universal Manipulation Interface, \cite{chi2024universal,ha2024umilegs}) systems use \textbf{one-way} communication by sending high-level policy outputs to low-level controllers via end-effector trajectories—often suboptimal or even infeasible for a given embodiment. Our approach introduces \textbf{two-way} communication, letting the low-level controller steer UMI policies from actions with \textbf{\textcolor{red}{high tracking cost}} to those with \textbf{\textcolor{custom_green_1}{lower cost}}, enabling more robust and high-performance cross-embodiment deployment.} 
    \vspace{-6mm}
    \label{fig:teaser}
\end{strip}

\begin{abstract}

We introduce UMI-on-Air, a framework for embodiment-aware deployment of embodiment-agnostic manipulation policies. Our approach leverages diverse, unconstrained human demonstrations collected with a handheld gripper (UMI) to train generalizable visuomotor policies. A central challenge in transferring these policies to constrained robotic embodiments—such as aerial manipulators—is the mismatch in control and robot dynamics, which often leads to out-of-distribution behaviors and poor execution. To address this, we propose Embodiment-Aware Diffusion Policy (EADP), which couples a high-level UMI policy with a low-level embodiment-specific controller at inference time. By integrating gradient feedback from the controller's tracking cost into the diffusion sampling process, our method steers trajectory generation towards dynamically feasible modes tailored to the deployment embodiment. This enables plug-and-play, embodiment-aware trajectory \textit{adaptation at test time}. We validate our approach on multiple long-horizon and high-precision aerial manipulation tasks, showing improved success rates, efficiency, and robustness under disturbances compared to unguided diffusion baselines.
Finally, we demonstrate deployment in previously unseen environments, using UMI demonstrations collected in the wild, highlighting a practical pathway for scaling generalizable manipulation skills across diverse—and even highly constrained—embodiments.
All code, data, checkpoints, and result videos can be found at \ourwebsite.

\end{abstract}

\section{Introduction}

There is a growing interest in extending manipulation beyond the lab and into more complex, dynamic settings. Among emerging embodiments, unmanned aerial manipulators (UAMs) hold particular promise. Essentially a manipulator with practically limitless reach, UAMs can access environments that are otherwise unreachable or unsafe, such as performing infrastructure maintenance atop towers or harvesting crops in cluttered orchards. Multiple research works have focused on those practical tasks, including non-destructive testing~\cite{bodie2019omnidirectional}, painting~\cite{vempati2018paintcopter}, drilling~\cite{ding2021design}, light bulb installation~\cite{he2025flying}, etc. These applications highlight the potential of UAMs as an embodiment, but scalable visuomotor policy learning for UAMs remains limited. 



A major bottleneck is data collection. Teleoperation is particularly challenging for UAMs due to expensive and fragile hardware and unintuitive interface. To address this, recent work explores cross-embodiment collection, with the Universal Manipulation Interface (UMI)~\cite{chi2024universal} offering a portable way to record demonstrations across environments. By decoupling demonstrations from specific robots, UMI enables training of embodiment-agnostic manipulation policies.

\begin{figure*}[t] 
    \centering
    \includegraphics[width=\linewidth]{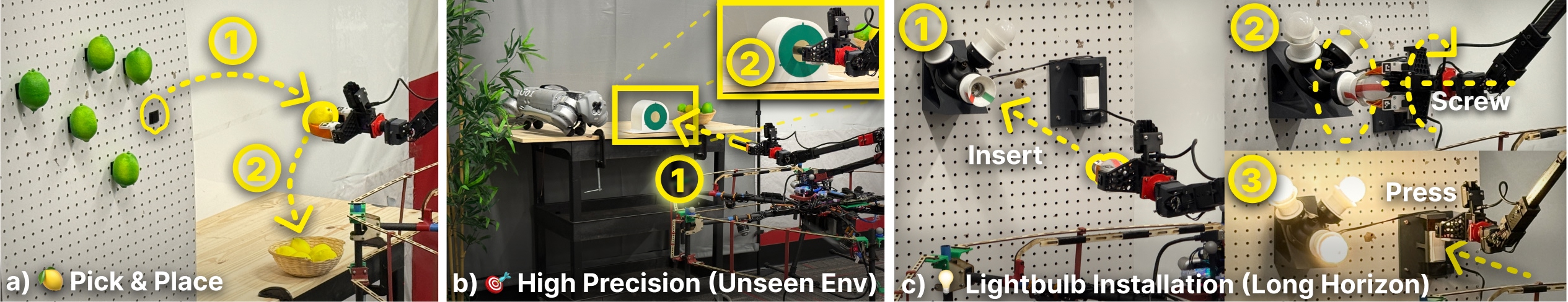}
    \caption{
    \textbf{Aerial Manipulation Tasks.}
    Combining UMI and our embodiment-aware guidance approach enables scalable data-collection and robust deployment of fully-autonomous skills previously beyond reach.
    On our UAM, we showcase (a) lemon harvesting (must find ripe yellow ones), (b) high precision peg insertion in \textit{unseen} environments, and (c) long-horizon light bulb installation tasks.
    }
    \vspace{-8mm}
    \label{fig:real_world_tasks}
\end{figure*}


While UMI enables embodiment-agnostic manipulation policies, their success hinges on the embodiment’s ability to execute the generated trajectories. Fixed-base arms with precise controllers are highly ``UMI-able\footnote{We will formally define ``UMI-able'' in \S~\ref{sec:experiment}, and Fig.~\ref{fig:simulation_results} quantifies how ``UMI-able'' different embodiments are in simulation.}'', able to execute UMI policies as if they were the handheld gripper. In contrast, embodiments like UAMs face stringent physical and control constraints such as stability under aerodynamic disturbances~\cite{shi2019neural,o2022neural}. Without accounting for these constraints, UMI policies may yield trajectories that are infeasible, unsafe, or inefficient. Hence, the \textbf{central challenge} is then how to extend UMI beyond highly UMI-able robots to embodiments where control constraints fundamentally shape feasibility.

To address this challenge, we propose \textbf{Embodiment-Aware Diffusion Policy (EADP)}, where the key idea is to enable \textbf{two-way} communication between an embodiment-agnostic high-level manipulation policy and embodiment-specific low-level controllers (Fig.~\ref{fig:teaser}). Unlike standard UMI systems~\cite{chi2024universal,ha2024umilegs} that rely on \textbf{one-way} communication by passing policy outputs directly to controllers, EADP lets low-level controllers guide the high-level policy through the denoising process, therefore producing end-effector (EE) trajectories that are more feasible for the target embodiment (e.g., a UAM).

Concretely, at each denoising step, the embodiment’s controller evaluates the noisy EE trajectory with a tracking cost, measuring its feasibility under current constraints. By backpropagating this cost to the noisy action trajectory, the policy is guided toward action trajectories that are more feasible. By leveraging the multi-modality of UMI policies (from diverse human data and the diffusion architecture), EADP biases the action generation toward strategies best aligned with the embodiment’s capabilities.

In summary, the work has three main contributions:
\begin{itemize}[leftmargin=4mm]
    \item We propose Embodiment-Aware Diffusion Policy, a framework that integrates embodiment-specific controller feedback into high-level trajectory generation by diffusion policies, enabling plug-and-play embodiment-aware trajectory guidance at test time.
    \item We introduce a simulation-based benchmark suite\footnote{All code, data, checkpoints, and result videos can be found at \ourwebsite}, which facilitates systematic investigation of the embodiment gap when using UMI demonstration data on embodiments with varying UMI-abilities. 
    \item We present UMI-on-Air, a system that validates our method on challenging aerial manipulation tasks (Fig.~\ref{fig:real_world_tasks}), outperforming embodiment-agnostic baselines.
\end{itemize}

By closing the gap between embodiment-agnostic policies and embodiment-specific constraints, this work is a step towards making all robots more UMI-able, thus extending scalable, universal manipulation skills to robots and environments previously beyond reach.

\section{Related Works}
\begin{figure*}[t]
    \centering
    \includegraphics[width=\linewidth]{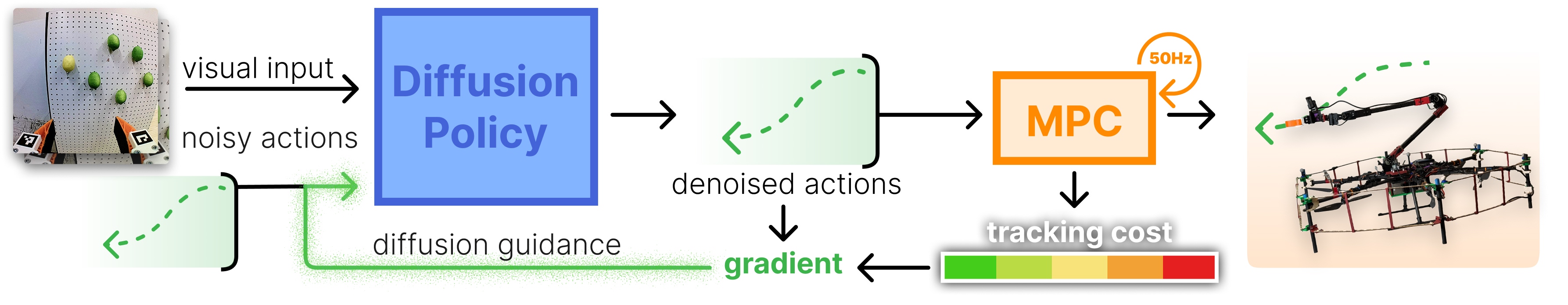}
    \caption{\textbf{Embodiment-Aware Diffusion Policy.}
    Using UMI, we collect data for an embodiment-agnostic \textbf{\textcolor{custom_blue_1}{Diffusion Policy}}, which iteratively denoises actions from visual inputs.
    To produce more feasible actions, we add \textbf{\textcolor{custom_green_1}{gradients}} of the MPC's tracking cost to the diffusion model's output at each iteration, steering the denoising process akin to classifier guidance.
    Finally, the guided action sequence is tracked by \textbf{\textcolor{orange}{MPC}} at 50Hz.
    }
    \label{fig:pipeline}
    \vspace{-7mm}
\end{figure*}

\subsection{Mobile Manipulation}
\textbf{Ground-Based Manipulation.} 
Ground-based mobile manipulation systems traditionally emerged to be designed to tailor to specific use cases~\cite{chestnutt2005footstep, feng2014optimization, krotkov2018darpa, bajracharya2024demonstrating}. This led to a strong reliance on task and motion planning and model-based control, to capture the unique embodiment kinematics and dynamics for the specific mobile system. Recent learning-based systems have shown success in leveraging behavior cloning~\cite{fu2024mobile, yang2024equibot, ze2024generalizable, he2024omnih2o}, reinforcement learning (RL) \cite{xia2021relmogen, hu2023causal, mendonca2024continuously}, combining reinforcement learning for locomotion and behavior cloning for manipulation~\cite{liu2024visual, ha2024umilegs} as an alternative for ground-based mobile manipulation via imitation learning. 


\textbf{Aerial Manipulation.} While aerial manipulation is a subset of mobile manipulation, it introduces distinct challenges compared to ground-based mobile embodiments, including disturbance near ground and wall, stability requirements, underactuated nonlinear dynamics, and strict payload constraints. Aerial manipulation has been demonstrated across a wide variety of applications, including inspection~\cite{bodie2019omnidirectional, guo2024aerial}, writing and painting~\cite{lanegger2022aerial, guo2024flying}, object grasping~\cite{ubellacker2024softgrasp, bauer2024an}, pick-and-place~\cite{cao2025proximal}, insertion~\cite{wang2023millimeter}, and articulated object interaction~\cite{brunner2022planning, lee2021manip}. These successes have typically relied on specialized hardware systems coupled with carefully engineered control strategies for specific tasks, but remain hard to scale to novel manipulation goals or environments. Recent research has shifted toward general frameworks that abstract away embodiment-specific dynamics, such as the EE-centric control interface that decouples high-level decision making from low-level embodiment-specific actuation~\cite{he2025flying}. However, robust and generalizable policies for such systems require large-scale \emph{robot} data that span diverse objects, scenes, and flight conditions. However, data collection directly with UAMs is challenging due to the difficulty of deploying drones across diverse physical environments, motivating alternative strategies for data collection and deployment.


\vspace{-1mm}

\subsection{Cross-embodiment Learning}
Recent works have explored large-scale cross-embodiment pretraining, involving data collected from various robotic embodiments and subsequently finetuned to accommodate specific hardware embodiments~\cite{liu2024rdt, yang2024pushinglimitscrossembodimentlearning, reed2022a,qiu2025humanoid}. 
These approaches rely on the assumption of a unified action space, which allows data sharing across robots with similar morphology, primarily robotic arms or mobile manipulators. Despite improvements in generalization, such methods require extensive embodiment-specific finetuning datasets to ensure policies adapt to the target embodiment.

An alternative strategy collected demonstrations directly from the human-embodiment using intuitive handheld interfaces~\cite{shafiullah2023bringing, song2020grasping, young2021visual, xu2025dexumi}. For instance, UMI~\cite{chi2024universal} bridges the embodiment gap by aligning the observation space and action spaces of the hand-held gripper and the robot embodiment, enabling large-scale, in-the-wild demonstrations without relying on physical robot hardware. However, policies trained directly from human demonstrations internalize action constraints reflective of human embodiment. As a result, these policies remain unaware of the distinct dynamics and physical limitations of embodiments like mobile manipulators~\cite{ha2024umilegs, he2025flying}, where the EE cannot precisely track the generated action sequences, leading to unreliable execution. 

To incorporate embodiment information into policy representations, recent works developed specialized model architectures. Embodiment-aware policies leveraged graph neural networks (GNNs), explicitly modeling robot structures as graphs, with joints as nodes and links as edges~\cite{wang2018nervenet, huang2020one}. 
Subsequent works explored transformer-based models, motivated by their superior representational capacity~\cite{kurin2020my, gupta2022metamorph, sferrazza2024body, patel2024get}. Embodiment-aware architectures showed impressive zero-shot generalization capabilities within RL contexts by using large-scale training with extensive embodiment randomization, yet their adoption within imitation learning remains limited due to data scarcity. In contrast, our work proposes incorporating embodiment-awareness during inference time by integrating feedback from a low-level embodiment-specific controller into a diffusion policy's trajectory generation process. By iteratively guiding diffusion-based sampling toward controller-feasible trajectories, we gain the benefits of abstraction from an ee-centric action space while producing trajectories that respect the robot's physical constraints.

\vspace{-1mm}

\section{Method}

\begin{figure*}[t] 
    \centering
    \includegraphics[width=\linewidth]{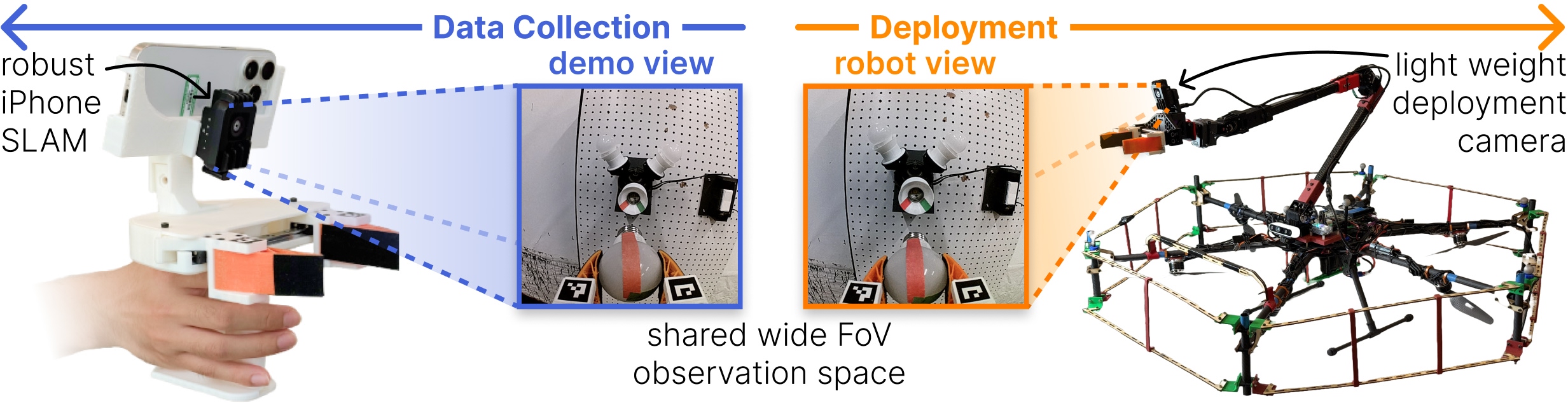}
    \caption{
    \textbf{Data Collection to Deployment.}
    Our data collection setup contains an iPhone running SLAM tracking, a lightweight camera for deployment, and compliant, 3D-printed gripper fingers. By sharing the observation and action space between data collection and deployment time, we minimize the embodiment gap.
    }
    \label{fig:data_collection}
    \vspace{-7mm}
\end{figure*}

\subsection{Data Collection Interface}

We adopt the UMI~\cite{chi2024universal} paradigm for human demonstration collection: a lightweight, hand-held gripper with a wrist-mounted camera for egocentric observation, and a shared action interface expressed in the EE frame. This design enables in-the-wild data collection without requiring robot hardware, and aligns the training and deployment modalities by replicating the camera–gripper configuration on the robot.

In our configuration (Fig. \ref{fig:data_collection}), we make three key modifications to the original UMI for UAM deployment. First, we replace the GoPro with a lightweight OAK-1 W camera, which reduces payload while maintaining a wide field of view. Second, we downsized the finger geometry to reduce the inertia of the EE. Finally, we use an iPhone-based visual–inertial SLAM system to more accurately track the 6-DoF EE pose during data collection. 

Each demonstration consists of synchronized egocentric RGB images, 6-DoF EE pose trajectories, and continuous gripper width tracked using fiducial markers on the fingers. These sequences form input--output pairs for policy learning: the input is an observation window consisting of images, relative EE poses, and gripper widths, while the output is a horizon of future actions given as relative EE trajectories and gripper widths. A conditional UNet-based~\cite{ronneberger2015u} diffusion policy is trained on these pairs, enabling the generation of multimodal action sequences from the UMI demonstrations.

\subsection{End-Effector-Centric Controllers}
\label{sec:controllers}

A key requirement for deploying embodiment-agnostic policies is a controller that interprets task-space reference trajectories $a = \{\bm{p}_{t}^r, \bm{R}_{t}^r\}_{t=1}^H$—a sequence of desired EE positions $\bm{p}^r \in \mathbb{R}^3$ and orientations $\bm{R}^r \in SO(3)$ over horizon $H$—into embodiment-specific actions. 
We adopt an \textit{EE–centric} perspective: the high-level policy always produces EE reference trajectories, while the controller is responsible for realizing them subject to embodiment constraints. 
This abstraction supports a spectrum of controllers, ranging from simple inverse kinematics (IK) with velocity limits to full model predictive control (MPC).

To guide the diffusion policy toward embodiment-feasible behaviors, we define a tracking cost $L_{\text{track}}(a)$ that evaluates how well a given trajectory $a$ can be executed by a particular controller. High tracking cost indicates trajectory segments that are difficult to follow due to dynamic infeasibility, underactuation, or control saturation, while low cost reflects better alignment with the embodiment capabilities.



\paragraph{Inverse Kinematics with Velocity Limits}  
For table-top manipulators and other robots with relatively simple dynamics, a lightweight controller can model the system well. At each step, the desired waypoint $(\bm{p}^r, \bm{R}^r)$ is mapped to a robot configuration $\bm{q}\in\mathbb{R}^{n}$—which may include both mobile base pose and arm joint angles—using the inverse kinematics function $\bm{f}_{\text{IK}}$, which maps a desired EE pose $(\bm{p}^r, \bm{R}^r)$ together with the current configuration $\bm{q}_t$ to a feasible robot configuration.  
We denote the per-step velocity bound as $\bm{\delta}_{\max} = \bm{\dot{q}}_{\max}\Delta t$, which accounts for the hardware velocity limit $\bm{\dot{q}}_{\max}$ and the controller timestep $\Delta t$.  
The forward kinematics $\bm{f}_{\text{FK}}(\bm{q})$ reconstructs the trajectory waypoint, and the tracking cost is the squared error between reconstructed and reference trajectories:

\begin{equation}
\bm{q}_{t+1} = \bm{q}_{t} + 
\text{clip }\!\Big( \bm{f}_{\text{IK}}(a_t, \bm{q}_t) - \bm{q}_{t},\;
-\bm{\delta}_{\max},\;
\bm{\delta}_{\max} \Big)
\label{eq:ik_update}
\end{equation}
\begin{equation}
L_{\text{track}}(a) = \sum_{t=1}^H \left\| \bm{f}_{\text{FK}}(\bm{q}_t) - a_t \right\|^2 
\label{eq:l_track_ik}
\end{equation}

This provides a feasible sequence of configurations and a differentiable tracking cost $L_{\text{track}}$.


\paragraph{Model Predictive Controller}

Richer controller instantiations can be used for robots that require accurate modeling of dynamics, such as UAMs.
We adopt the EE–centric whole-body MPC from~\cite{he2025flying}, including the formulation and all parameters. The MPC is solved using ACADOS~\cite{verschueren2022acados}. This controller coordinates UAV and manipulator motion by optimizing a finite-horizon cost function subject to dynamics and actuation constraints. The state and control variables are defined as:
\begin{align}
\bm{x} := 
\begin{bmatrix}
\bm{p} &
\bm{R} &
\bm{v} &
\bm{\theta}
\end{bmatrix}, \quad
\bm{u} := 
\begin{bmatrix}
\bm{\tau} &
\bm{\theta}_{\text{cmd}}
\end{bmatrix},
\end{align}

where $\bm{v} \in \mathbb{R}^6$ is the body velocity (linear + angular), $\bm{\theta}\in\mathbb{R}^{n}$ are the manipulator’s joint angles, $\bm{\tau} \in \mathbb{R}^6$ is the commanded wrench (forces and torques) and $\bm{\theta}_{\text{cmd}}\in\mathbb{R}^{n}$ are the commanded joint angles. Note that we adopt a similar UAM system with~\cite{he2025flying}, which is a fully-actuated hexarotor, allowing us to send the commanded 6-dim control wrench directly. 

The cost functions are defined in terms of errors between the predicted and reference values:
\begin{subequations}
\begin{align}
\bm{e}_p &= \bm{p} - \bm{p}^r \\
\bm{e}_R &= \frac{1}{2} \left( \bm{R}^r{}^\top \bm{R} - \bm{R}^\top \bm{R}^r \right)^\vee \\
\bm{e}_v &= \bm{v} - \bm{v}^r \\
\bm{e}_\theta &= \bm{\theta} - \bm{\theta}^r \\
\bm{e}_u &= \bm{u} - \bm{u}^r
\end{align}
\end{subequations}
where $(\cdot)^r$ denotes reference values, and $(\cdot)^\vee$ denotes the vee-operator that maps a skew-symmetric matrix to $\mathbb{R}^3$. The default reference joint angles $\bm{\theta}^r$ are pre-defined, the reference velocity is set to $\bm{v}^r = [\bm{0}_6]$ and the reference control $\bm{u}^r = [\bm{0}_6, \hat{\bm{\theta}}]$ assumes zero wrench and current joint positions $\hat{\bm{\theta}}\in\mathbb{R}^{n}$. 

The optimal control sequence is obtained by solving the following finite-horizon constrained optimization: 
\begin{subequations}
\begin{align}
\bm{u}_{\text{opt}} = \arg\min_{\bm{u}} \; & \left\{ 
L_e(\bm{x}_H, \bm{x}_H^r) + \sum_{t=1}^{H-1} L_r(\bm{x}_t, \bm{x}_t^r, \bm{u}_t) \right\} \label{eq:mpc_cost} \\
\text{s.t.} \quad & \bm{x}_{t+1} = \bm{f}_{\text{dyn}}(\bm{x}_t, \bm{u}_t) \label{eq:mpc_dyn} \\
& \bm{x}_0 = \hat{\bm{x}}, \quad \bm{x}_t \in \mathcal{X} \label{eq:mpc_state} \\
& \bm{u}_{\text{lb}} \leq \bm{u}_t \leq \bm{u}_{\text{ub}} \label{eq:mpc_input}
\end{align}
\end{subequations}
where $\bm{f}_{\text{dyn}}$ is the system dynamics, $\mathcal{X}$ the feasible state space, and $\bm{u}_{\text{lb}}, \bm{u}_{\text{ub}}$ the actuation bounds. 
Eq. (\ref{eq:mpc_cost}) uses terminal cost $L_e$ and stage cost $L_r$ that are quadratic functions of the errors, given by $
\bm{e}^\top \bm{Q} \bm{e}, \quad \bm{e} \in \{\bm{e}_p, \bm{e}_R, \bm{e}_v, \bm{e}_\theta, \bm{e}_u\}
$
where $\bm{Q}$ matrices are hand-tuned positive definite weights. 
Discretization is performed using a fourth-order Runge–Kutta scheme for stability.

In addition to producing control inputs, the MPC exposes a tracking cost $L_{\text{track}}$ that quantifies how well the reference trajectory $a$ can be followed under these constraints:

\begin{equation}
L_{\text{track}}(a) = \sum_{t=1}^H \left( 
\bm{e}_{p,t}^\top \bm{Q}_p \bm{e}_{p,t} + 
\bm{e}_{R,t}^\top \bm{Q}_R \bm{e}_{R,t}
\right)
\label{eq:track}
\end{equation}

\subsection{Embodiment-Aware Diffusion Guidance}
\label{sec:mpc_guided_diffusion}

We compute the gradient of $L_{\text{track}}$ with respect to the reference trajectory, \(\nabla_{a} L_{\text{track}}(a)\), which captures how sensitive the tracking error is to changes in the reference trajectory. In other words, it tells us how to nudge the reference trajectory $a$ so that it becomes more trackable by the low-level controller.

As illustrated in Fig.~\ref{fig:pipeline}, at inference time, we guide the conditional diffusion policy using this gradient feedback from the low-level controller. Let $a^k$ denote the noisy reference trajectory sample at diffusion timestep $k \in \{K, \dots, 1\}$, conditioned on observation data $\bm{o}$. We use the standard DDIM~\cite{song2020denoising} update step, given by:
\begin{equation}
a^{k-1} = a^k + \psi_k(\pi_\theta(a^k, t \mid \bm{o})),
\end{equation}
where $\pi_\theta$ is the trained denoiser, and $\psi_k$ is the DDIM update function for step $k$.

We incorporate gradient feedback from the tracking cost $L_{\text{track}}$ defined in Eq.~(\ref{eq:l_track_ik}, \ref{eq:track}) similarly to classifier-based guidance~\cite{dhariwal2021diffusion}. Specifically, we apply a guidance step to the trajectory sample toward feasible modes:
\begin{equation}
\tilde{a}^k = a^k - \lambda \cdot \bar{\omega}_k \cdot \nabla_{a^k} L_{\text{track}}(a^k),
\end{equation}
where $\lambda$ is a global guidance scale, and $\bar{\omega}_k \in (0,1)$ is the guidance scheduler, equal to the cumulative noise schedule $\bar{\alpha}_k$. This makes guidance scale time-dependent: weak during early noisy steps and stronger during later denoising steps. We then use the nudged sample for denoising.

\begin{algorithm}[h]
\caption{Embodiment-Aware DDIM Sampling}
\label{alg:mpc_guided_ddim}
\begin{algorithmic}[1]
\State Initialize $a^K \sim \mathcal{N}(0, I)$ \Comment{Start from noise}
\For{$k = K, \dots, 1$}
    \State \textcolor{custom_green_1}{$\tilde{a}^k \gets a^k - \lambda \cdot \bar{\omega}_k \cdot \nabla_{a^k} L_{\text{track}}(a^k)$}
    \State $a^{k-1} \gets \tilde{a}^k + \psi_k(\pi_\theta(\tilde{a}^k, k \mid \bm{o}))$
\EndFor
\State \Return $a^0$ \Comment{Reference trajectory}
\end{algorithmic}
\end{algorithm}

The full procedure is summarized in Algorithm~\ref{alg:mpc_guided_ddim}. The diffusion policy training remains independent of the embodiments used for deployment, but embodiment-specific controllers can inject real-time constraints and feasibility gradients.
Thus, our method robustifies plug-and-play deployment across embodiments without retraining. 

\vspace{-1mm}

\section{Experimental Results}
\label{sec:experiment}

Our experiments aim to evaluate the extent to which embodiment-aware guidance improves the deployment of embodiment-agnostic visuomotor policies. We design both simulation and real-world studies to probe the embodiment gap and assess how well EADP addresses it. The key questions we investigate are the following:

\begin{enumerate}[leftmargin=6mm]
    \item How significant is the embodiment gap across different robots, and to what extent does EADP mitigate it?  
    \item Does EADP enable reliable transfer of UMI-trained policies to real-world UAMs?  
    \item Can UMI-on-Air generalize to unseen environments?  
\end{enumerate}

\subsection{Simulation Experiments}

\begin{figure*}[t] 
    \centering
    \includegraphics[width=\linewidth]{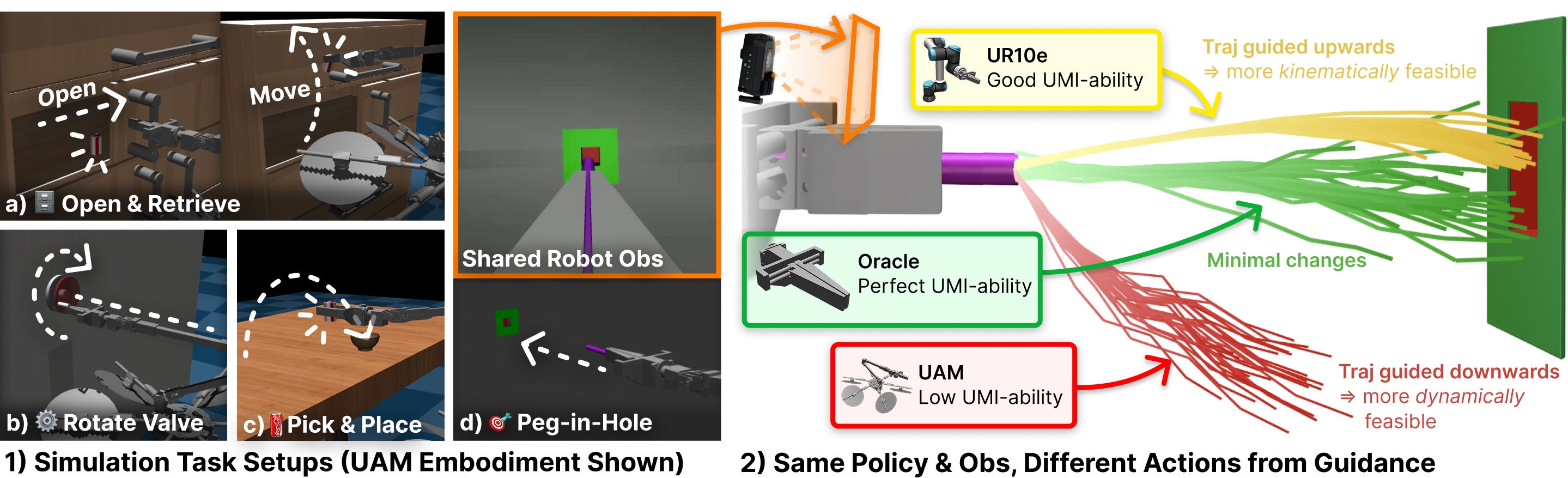}
    \caption{
    \textbf{Policy Adaptation Across Embodiments.}
    Across four simulated tasks (1) and three embodiments (2), we observe that EADP can adapt the embodiment-agnostic diffusion policy to the deployment embodiments with varying ``UMI-abilities''.
    Visualizing 32 action samples across different embodiments for the same observation, we observe that \textbf{UR10e}'s trajectories is guided upwards to be \textcolor{custom_yellow_1}{\textbf{more kinematically feasible}}, avoiding kinematic singularities.
    In contrast, the \textbf{UAM}'s trajectories are guided downwards to be \textcolor{red}{\textbf{more dynamically feasible}} due to perturbations along the $-Z$ direction.
    }
    \label{fig:task_setup}
    \vspace{-6mm}
\end{figure*}

To address our first question, we construct a controlled simulation benchmark in MuJoCo. Our setup allows us to systematically evaluate how an embodiment can affect policy execution across tasks.

We use motion capture on a UMI gripper to collect human demonstrations in simulation, mirroring the handheld demonstration process used in the real world. These demonstrations are used to train an \textit{embodiment-agnostic} Diffusion Policy (DP), which serves as the base policy for comparison with EADP. We evaluate across four simulation environments, covering both long-horizon and precision tasks:  
\begin{enumerate}[leftmargin=6mm]
    \item \textit{Open-And-Retrieve}: Slide open a cabinet, pick up can, and place on top of cabinet. Can location is randomized.
    \item \textit{Peg-In-Hole}: Insert a 1cm peg into a 2cm square hole. 50s timeout if not pegged. Hole location is randomized.
    \item \textit{Rotate-Valve}: Rotating a valve to a specified orientation.  Valve location is randomized. 
    \item \textit{Pick-and-Place}: Lift a can and place it in a bowl. Object locations are randomized.
\end{enumerate}

We deploy the trained policy across three embodiments, each reflecting different levels of control fidelity:  
\begin{enumerate}[leftmargin=6mm]
    \item \textit{Oracle}: A flying gripper that perfectly tracks the policy-generated trajectory. This provides an upper bound on achievable performance with no embodiment gap.  
    \item \textit{UR10e}: A fixed-base 6-DoF manipulator, using an IK-based velocity-limited controller (\S~\ref{sec:controllers}).
    \item \textit{UAM}: Aerial manipulator using the MPC controller (\S~\ref{sec:controllers}). We consider two variants: (i) \textit{UAM (no disturbance)}, and (ii) \textit{UAM+Disturbance}, where we inject noise into the UAM base to simulate the $\sim$3\,cm average tracking error observed on hardware when hovering near a still target. This allows us to test whether EADP can help compensate for real-world disturbances.  
\end{enumerate}

\begin{figure*}[ht] 
    \centering
    \includegraphics[width=\linewidth]{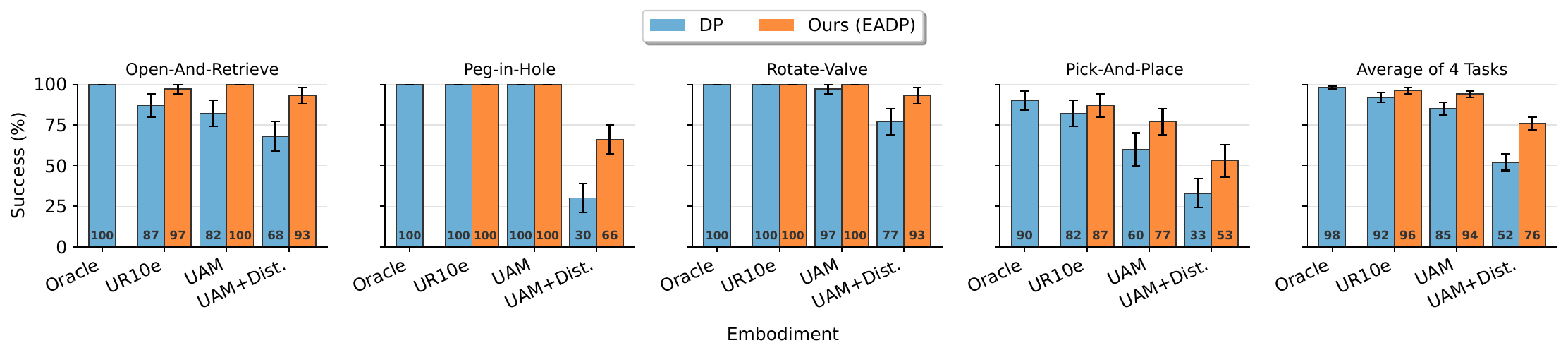}
    \vspace{-6mm}
    \caption{
    \textbf{Simulation Results}
    for Diffusion Policy (DP) and Embodiment-Aware Diffusion Policy (EADP) on four tasks across four embodiments. EADP consistently outperforms DP, with larger gains in more constrained (less ``UMI-able'') embodiments.
    }
    \label{fig:simulation_results}
    \vspace{-4mm}
\end{figure*}

Fig.~\ref{fig:simulation_results} reports success rates of DP and EADP across all tasks and embodiments. The gap between the \textit{Oracle} and baseline DP serves as a natural measure of how ``UMI-able'' each embodiment is. As expected, the \textit{UR10e} is close to \textit{Oracle} performance, reflecting that tabletop manipulators with IK controllers can reliably track UMI policies. In contrast, the \textit{UAM} exhibits a much larger gap—especially under disturbances—highlighting the difficulty of executing embodiment-agnostic trajectories on aerial systems.

EADP consistently reduces this embodiment gap. For \textit{UR10e}, improvements are modest but noticeable on difficult tasks. For the \textit{UAM}, EADP substantially boosts performance, recovering over 9\% on average without disturbances and over 20\% with disturbances. Even in the most constrained setting, EADP narrows the gap toward \textit{Oracle}, confirming that embodiment-aware guidance enables policies to adapt trajectories to dynamic feasibility.

The \textit{Open-and-Retrieve} task illustrates the challenges of long-horizon execution. Failures often occur when the gripper jams on the cabinet door or when placing the can on top—\textit{UR10e} slows down near its kinematic limits, while the \textit{UAM} overshoots with momentum, causing collisions. Disturbances exacerbate these issues, pushing trajectories out-of-distribution (OOD). EADP mitigates many of these cases by steering trajectories toward safer, more in-distribution motions given by the policy.

In the \textit{Peg-in-Hole} task, all embodiments succeed except the UAM with disturbances, where the hole is smaller than the average noise. This makes the task a stress test for disturbance robustness. EADP substantially improves reliability here, effectively rejecting infeasible pegging attempts under high noise and timing insertions when feasible, demonstrating that embodiment-aware guidance can even correct precision-sensitive behaviors (See Fig.~\ref{fig:task_setup}).

\begin{figure}[t] 
    \centering
    \includegraphics[width=\linewidth]{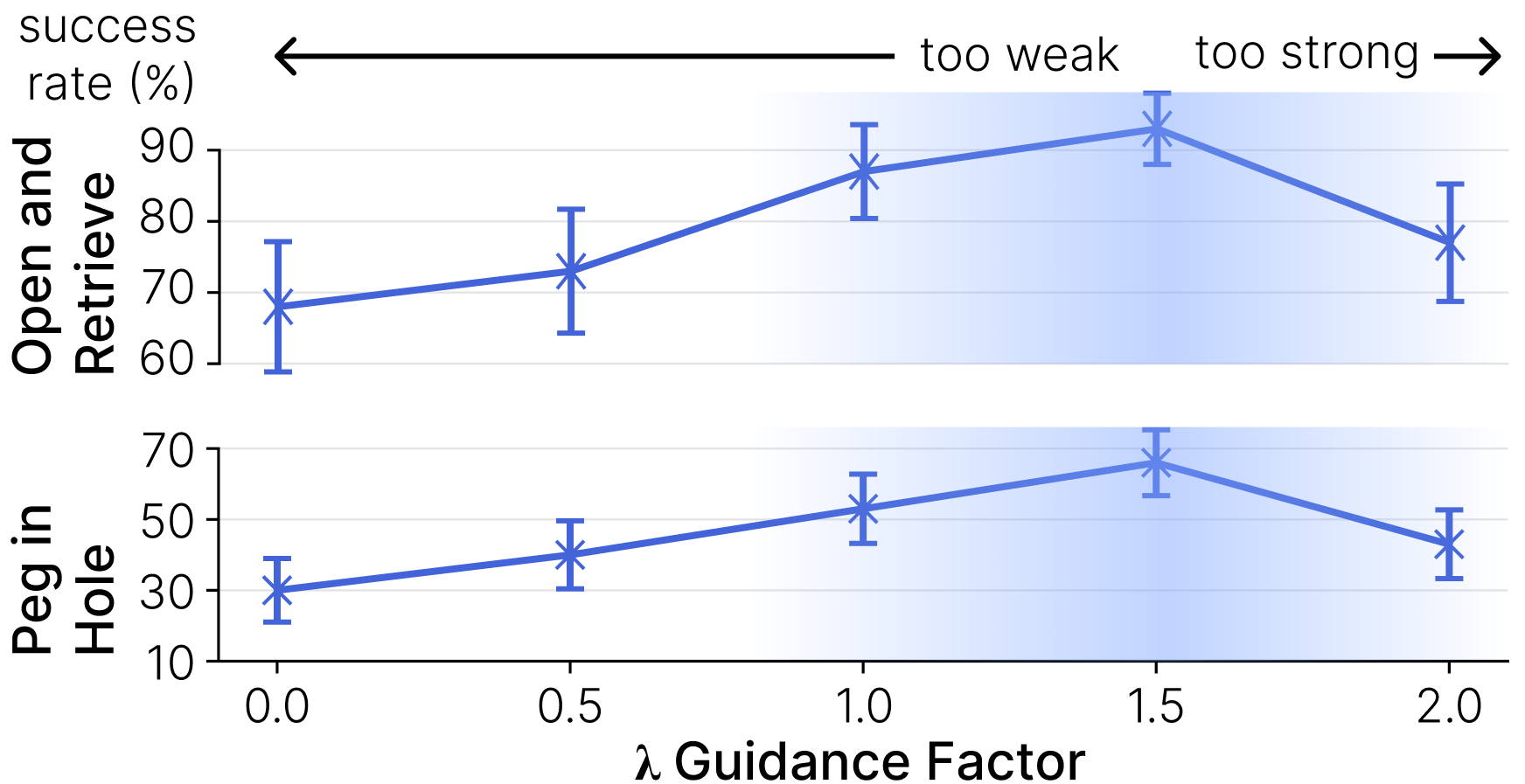}
    \vspace{-4mm}
    \caption{
    \textbf{Guidance Ablation} for UAM+Disturbance.
    }
    \label{fig:lambda_ablation}
    \vspace{-6mm}
\end{figure}

The global guidance scale $\lambda$ controls the trade-off between task-oriented trajectory generation and controller-feasible execution (Fig.~\ref{fig:lambda_ablation})
Without guidance ($\lambda=0$), performance collapses under disturbances. As $\lambda$ increases, success rates steadily improve. Excessively large $\lambda$ over-constrains the denoising process, leading to conservative, OOD behaviors.

\subsection{Real-world Experiments}

\begin{figure}
    \centering
    \includegraphics[width=0.99\linewidth]{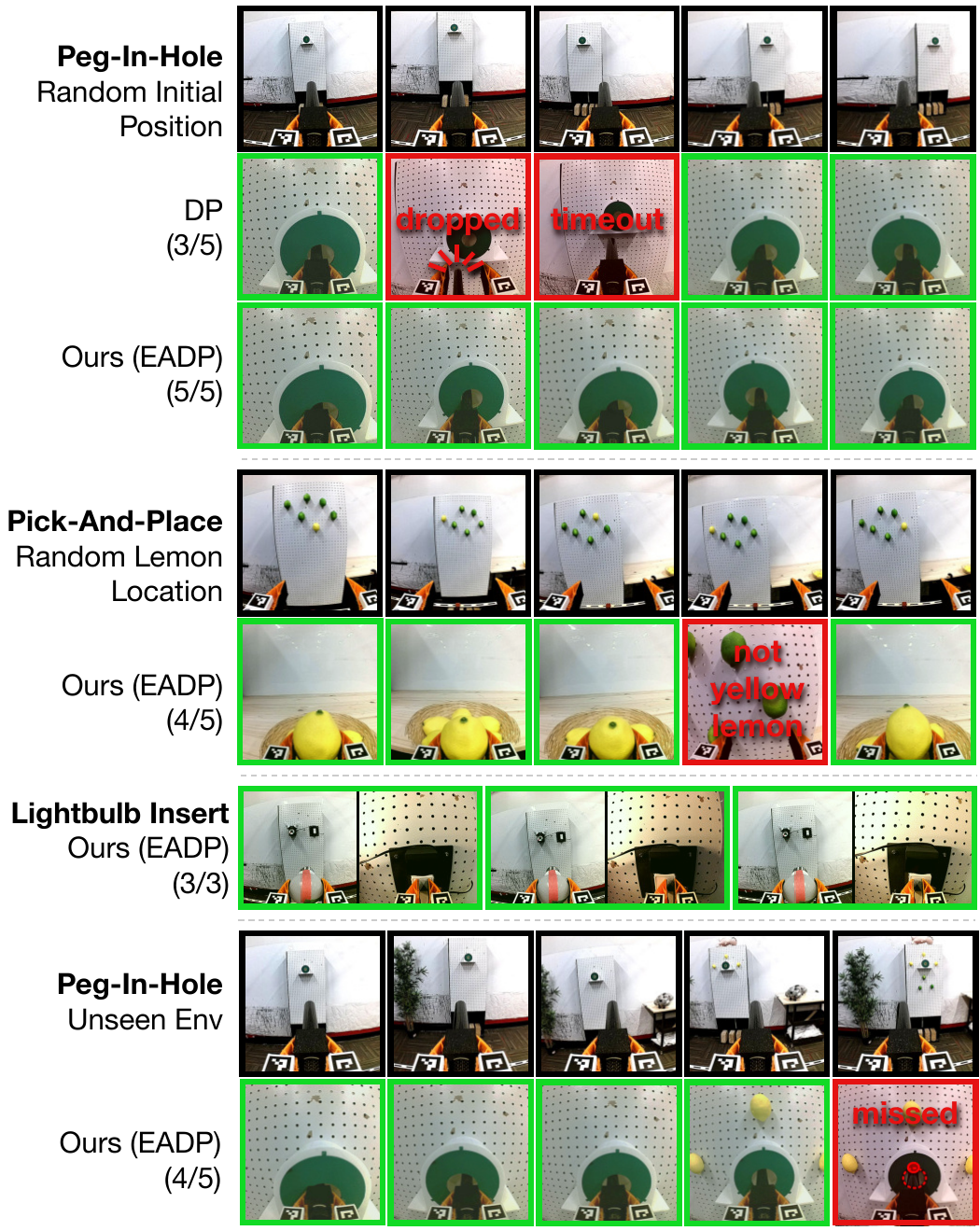}
    \caption{\textbf{Real-World Results} for DP and EADP.
    Colored borders indicate \textbf{\textcolor{custom_green_1}{success}} or \textbf{\textcolor{red}{failure}} for each trial.}
    \label{fig:experiment_all_image}
    \vspace{-6mm}
\end{figure}

We next address our second research question: \textit{does EADP enable reliable transfer of UMI-trained policies to real-world UAMs?} To this end, we evaluate on three aerial manipulation tasks (Fig.~\ref{fig:real_world_tasks}) that span precision, robustness, and long-horizon execution, followed by a cross-environment generalization test. 
We conducted experiments using a fully actuated hexarotor drone with a $4$ DoF manipulator and a gripper.
A motion capture system provides the drone state, while the EE state is computed using forward kinematics. Policy inference runs on an NVIDIA RTX 4070Ti at approximately 1.5\,Hz; the MPC executes on CPU at 50\,Hz, interpolating between asynchronously published reference waypoints.
Fig.~\ref{fig:experiment_all_image} summarizes results across all trials.

\paragraph{Peg-in-Hole.}  
We evaluate on a 4\,cm hole with a 2\,cm peg, with randomized starting positions and a 3\,min timeout. While the baseline DP failed due dropped peg or a timeout, EADP succeeded on all five trials (5/5). By incorporating controller feedback, EADP generated trajectories that avoided premature release and improved timing during insertion.

\paragraph{Pick-and-Place (Lemon Harvesting).}  
In this task, the UAM must harvest a lemon from a randomized location and place it into a basket. EADP completed 4/5 trials successfully, with the only failure occurring when an unripe (green) lemon was selected. Notably, in this failure case the ripe yellow lemon was positioned near the edge of the camera’s field of view from the initial state. Overall, EADP robustly handles aerial pick-and-place motions once a valid target is identified.

\paragraph{Lightbulb Insertion.}  
This long-horizon task requires threading a bulb into its socket until tight, followed by flipping the switch to confirm success. The task spans over 3 minutes of wall-clock time, underscoring the need for stability over extended horizons. EADP succeeded in all trials (3/3), demonstrating its ability to maintain precision and robustness throughout long-horizon tasks.

\paragraph{Cross-Environment Generalization.}  
Finally, we revisit the peg-in-hole task to probe our third research question: \textit{can UMI-on-Air generalize to previously unseen environments?} We collect a dataset of demonstrations in varied real-world settings distinct from the test environment, then evaluate in a new environment with gradually increasing distractions across trials. With a 5\,cm hole, EADP consistently located and aligned the peg, succeeding in 4/5 attempts. The only failure occurred when the drone collided with the hole’s enclosure, leading to a miss.

\section{Conclusion \& Discussion}

We introduced Embodiment-Aware Diffusion Policy (EADP), a framework for coupling embodiment-agnostic visuomotor policies with embodiment-specific controllers at inference time. Unlike standard UMI deployments that rely on one-way communication from high-level policy to low-level control, EADP enables controllers to provide gradient feedback on tracking feasibility, steering the diffusion sampling process toward trajectories that are dynamically feasible for the executing embodiment. This mechanism allows plug-and-play adaptation of UMI policies across diverse embodiments without retraining. Through large-scale simulation and real-world aerial manipulation experiments, we demonstrated that EADP consistently reduces the embodiment gap, especially in embodiments that are less ``UMI-able.''

While promising, our work leaves several directions for future research. First, the current system operates with a temporal gap between policy inference (around 1-2\,Hz) and high-frequency control (50\,Hz). This mismatch could be alleviated through streaming diffusion methods~\cite{jiang2025streaming} or continuous guidance mechanisms that allow tighter integration between policy and controller. Second, while we demonstrate EADP with IK and MPC instantiations, the framework is not limited to analytical controllers. It can be naturally extended to learned or reinforcement learning–based controllers using learned dynamics models.
We note that sampling-based guidance methods such as MPPI~\cite{williams2017model} offer an alternative to gradient-based steering, and achieve comparable results in simulation but are prohibitively expensive in practice due to repeated MPC evaluations on CPU.

By bridging the gap between general, data-driven visuomotor policies and embodiment-specific feasibility, EADP represents a step toward scalable, universal manipulation. In doing so, it expands the practical deployment of UMI-style demonstrations from controlled lab settings to robots and environments that were previously beyond reach.

\section{Acknowledgment}

This work was supported by the Robotics Institute Summer Scholars program. We thank Zeji Yi and Junyi Geng for their thoughtful discussions, and Yutong Wang and Mohammadreza Mousaei for their help with the experiments. This work was partially funded by NSF Award \#2512805, \#2143601, \#2037101, and \#2132519 and Toyota Research Institute. Guanya Shi holds concurrent appointments as an Assistant Professor at Carnegie Mellon University and as an Amazon Scholar. This paper describes work performed at Carnegie Mellon University and is not associated with Amazon. Dongjae Lee was supported by Basic Science Research Program through the National Research Foundation of Korea (NRF) funded by the Ministry of Education (RS-2025-02634317). 

\bibliographystyle{IEEEtran}
\bibliography{references}

\end{document}